\pdfoutput=1
\documentclass[conference]{IEEEtran}
\IEEEoverridecommandlockouts
\pdfoutput=1
\PassOptionsToPackage{table, x11names, dvipsnames, svgnames}{xcolor}
\usepackage{lipsum}
\usepackage{ dsfont }
\usepackage{cite}
\usepackage{bbm}
\usepackage{amsmath,amssymb,amsfonts}
\usepackage{algorithm, algorithmic}

\usepackage{float}
\makeatletter
\let\newfloat\newfloat@ltx
\makeatother

\usepackage{graphicx}
\usepackage{textcomp}
\usepackage{xcolor}
\usepackage[english]{babel}
\usepackage[acronym,nonumberlist]{glossaries}
\usepackage{glossaries-prefix}
\usepackage{pgfplots}
\usepgfplotslibrary{fillbetween}
\usepackage{amssymb}
\usepackage{import}
\usepackage{bm}
\usepackage{siunitx}
\usepackage{subcaption}
\usepackage{braket}
\usepackage{mathtools}

\usepackage{hyperref}
\hypersetup{%
linktocpage=true, 
colorlinks=false,
pdfborder={0 0 0}, 
breaklinks=true, pdfpagemode=UseNone, pageanchor=true, pdfpagemode=UseOutlines,%
plainpages=false, bookmarksnumbered, bookmarksopen=true, bookmarksopenlevel=1,%
hypertexnames=true, pdfhighlight=/O,
pdftitle={Fourier Analysis of Variational Quantum Circuits for Supervised Learning},%
pdfauthor={Marco Wiedmann},%
pdfsubject={QML},%
pdfkeywords={},%
pdfcreator={pdfLaTeX},%
pdfproducer={LaTeX with hyperref}%
}

\usepackage{pgfplots}
\pgfplotsset{width=7cm,compat=1.8}

\usepackage{multirow}
\usepackage{color}
\usepackage{tabularray}

\usepackage[inline]{enumitem}

\usepackage[capitalise]{cleveref}

\makeatletter
\providecommand{\@LN@col}[1]{}
\makeatother

\newacronym{ML}{ML}{machine learning}
\newacronym{MSE}{MSE}{mean squared error}
\newacronym{QML}{QML}{quantum machine learning}
\newacronym{NISQ}{NISQ}{noisy intermediate scale quantum}
\newacronym{QC}{QC}{quantum circuit}
\newacronym{VQC}{VQC}{variational quantum circuit}
\newacronym{VQA}{VQA}{variational quantum algorithm}
\newacronym{DRU}{DRU}{data re-uploading}
\newacronym{SPSA}{SPSA}{simultaneous perturbation stochastic approximation}
\newacronym{FFT}{FFT}{fast Fourier transform}
\newacronym{NFFT}{NFFT}{nonequispaced fast Fourier transform}

\usepackage{yquant}
\usepackage{braket}
\usetikzlibrary{quotes, fit}
\usepackage{environ}

\makeatletter
\newsavebox{\measure@tikzpicture}
\NewEnviron{scaletikzpicturetowidth}[1]{%
  \def\tikz@width{#1}%
  \begin{lrbox}{\measure@tikzpicture}%
  \BODY
  \end{lrbox}%
  \pgfmathparse{#1/\wd\measure@tikzpicture}%
  \BODY
}
\makeatother

\begin{document}

\title{Fourier Analysis of Variational Quantum Circuits for Supervised Learning
\thanks{
The research is supported by the Bavarian Ministry of Economic Affairs, Regional Development and Energy with funds from the Hightech Agenda Bayern via the project BayQS.\\
email address for correspondence: 
maniraman.periyasamy@iis.fraunhofer.de}
}

\author{
\IEEEauthorblockN{Marco Wiedmann, Maniraman Periyasamy, Daniel D. Scherer}
\IEEEauthorblockA{\textit{Fraunhofer IIS, Fraunhofer Institute for Integrated Circuits IIS},
Nuremberg, Germany \\\vspace{1mm}}
}

\maketitle

\begin{abstract} 
    \Glspl{VQC} can be understood through the lens of Fourier analysis. It is already well-known that the function space represented by any circuit architecture can be described through a truncated Fourier sum. We show that the spectrum available to that truncated Fourier sum is not entirely determined by the encoding gates of the circuit, since the variational part of the circuit can constrain certain coefficients to zero, effectively removing that frequency from the spectrum.
    To the best of our knowledge, we give the first description of the functional dependence of the Fourier coefficients on the variational parameters as trigonometric polynomials. This allows us to provide an algorithm which computes the exact spectrum of any given circuit and the corresponding Fourier coefficients. Finally, we demonstrate that by comparing the Fourier transform of the dataset to the available spectra, it is possible to predict which \gls{VQC} out of a given list of choices will be able to best fit the data.
\end{abstract}

\glsresetall
\section{Introduction}
\label{sec:introduction}


With \gls{NISQ} devices now being available from several providers, the question of what these devices might be capable of becomes ever more important.
\Glspl{VQA} have garnered a lot of attention in recent years as potentially demonstrating quantum utility in a plethora of fields like e.g. quantum chemistry \cite{peruzzo2014variational, kandala2017hardware, parrish2019quantum, li2019variational, zhang2021shallow}, 
\gls{QML} \cite{schuld2015introduction, biamonte2017quantum, schuld2020circuit, guan2021quantum, meyer2023quantum} 
and quantum reinforcement learning \cite{wu2020quantum, jerbi2021parametrized, meyer2022survey, cherrat2023quantum, periyasamy2023batch}.

In all of these cases the idea can be briefly summarized as follows:
A \gls{VQC} is used to prepare a parametrized quantum state \(\ket{\psi(\theta)}\). The expectation value \(\bra{\psi(\theta)} \mathcal{O} \ket{\psi(\theta)}\) of a given observable \(\mathcal{O}\) with respect to this state is estimated from several shots of measurements.
This expectation value is then fed into a cost function $L(\theta) = f\left(\langle \mathcal{O} \rangle_\theta\right)$, which is typically optimized by some classical black-box optimization algorithm.

In this paper, we want to focus on \glspl{VQA} in the context of supervised learning problems, where classical input data is encoded into the \gls{VQC} alongside the optimization parameters.
In this case the \gls{VQC} acts as a function approximator that ideally should be able to represent the classical data.
It has recently been established that the output of such a \gls{VQC} can be understood in terms of a finite Fourier sum of the inputs \cite{Schuld2021effect}.
Determining the frequencies that are present in this Fourier sum allows us to characterize the set of functions that can be realized by a given circuit architecture.
This information has been used to study the effects of various ansatz choices on the expressive power of the \gls{VQC}.

However, purely maximizing the expressivity of a \gls{VQC} is not a desirable goal.
The reason for this is an inherent tradeoff between expressivity and trainability in \glspl{VQA} \cite{holmes2022connecting}, i.e. the broader the class of functions that can be realized the harder it becomes to actually find the good candidates.
Therefore it would be much better if \glspl{VQC} could be specifically engineered for a given task.

The contributions of our paper are twofold:
Firstly, we propose an algorithm which fully characterizes the Fourier sum that represents any given \gls{VQC}.
Our algorithm heavily builds on the one described in \cite{Nemkov_2023} to extract both the exact set of frequencies and their coefficients.
We realize that the coefficients consist of a combinatorial part and a sum of trigonometric polynomials and derive necessary and sufficient conditions for any of the combinatorial parts to vanish.
Therefore we obtain a more detailed description of the spectrum as the one previously derived in \cite{Schuld2021effect}.
Strictly speaking the spectrum is overestimated by the formula given in \cite{Schuld2021effect}, as some of the coefficients might vanish (cf. \cref{sec:related_work} for more details).
To the best of our knowledge, this is the first algorithm capable of giving the exact functional relationship between the variational parameters and the Fourier coefficients.
Secondly, we present a dataset-aware method of selecting a \gls{VQC} architecture out of a set of proposals.
The idea is that one can extract the most important frequencies, i.e. the ones with the largest Fourier coefficients, out of the dataset and select the simplest \gls{VQC} whose spectrum contains these frequencies.
This method is tested with various architectures both on synthetic and real-world data.
However, we need to note that calculating  spectrum and Fourier coefficients of any given \gls{VQC} facilitates classical simulation and therefore necessarily scales exponentially for classically hard architectures.

The paper is structured as follows:
After \cref{sec:background} establishes the basic theory behind the Fourier decomposition of \glspl{VQC}, \cref{sec:related_work} gives a summary of related papers.
\Cref{sec:calculate_spectrum} explains how the frequency spectrum and the associated Fourier coefficients are calculated.
Next, \cref{sec:ranking} establishes how the frequency spectrum can be compared to the Fourier decomposition of the dataset in order to predict which \gls{VQC} will be able to fit the data best.
\Cref{sec:experimental_setup} details how the numerical experiments on the architecture selection were done and \cref{sec:results} presents the results thereof.
Finally, the paper concludes with a summary in \cref{sec:conclusion}.
\section{Theoretical background}%
\label{sec:background}\noindent%

\subsection{Variational Quantum Circuits as Machine Learning Models}
\Glspl{VQA} are a subclass of quantum algorithms that act as function approximators in the context of \gls{ML}.
The corresponding \gls{VQC} is comprised of three types of quantum gates.

Firstly, the so-called encoding gates are used to feed the classical data into the model.
They are implemented as parameterized gates with the gate parameter being dependent on the input data.
Pauli rotations are a typical choice for the encoding gates.
The input data \(x\) is usually normalized into the interval \([-\pi, \pi]\) or \([0, 2\pi]\) so it can be directly used as the rotation angle.
Secondly, the free parameters \(\theta\) of the parameterized variational gates are used in the classical optimization routine to fit the data to the desired output.
And lastly, entangling gates like the CNOT or CZ gates build up entanglement in the circuit.
This is necessary to give the \gls{VQA} access to functions that are hard to simulate on a classical computer.

The \gls{VQA} outputs the expectation value of a chosen observable \(\mathcal{O}\) with respect to the quantum state that is prepared by the gates in the \gls{VQC}.
If we denote the unitary realized by the \gls{VQC} on a \(n\)-qubit system by \(U(x, \theta)\), then the \gls{VQA} represents the function
\begin{equation}
	f_\theta(x) = \bra{0}^{\otimes n} U^\dagger(x, \theta) \mathcal{O} U(x, \theta) \ket{0}^{\otimes n}.
\end{equation}

\subsection{Fourier Expansion of Variational Quantum Circuits}
The Fourier expansion is a useful tool to characterize the class of functions that any given \gls{VQA} can represent.
The output of a \gls{VQA} can be written as a partial Fourier series
\begin{equation}
	f_\theta(x) = \sum_{\omega \in \Omega} c_\omega(\theta) e^{i\omega x}
\end{equation}
with spectrum \(\Omega\) and coefficients \(c_\omega(\theta)\) \cite{Schuld2021effect}.

While the spectrum \(\Omega\) is determined purely by the number and choice of encoding gates, the Fourier coefficient \(c_\omega(\theta)\) of each frequency \(\omega\) only depends on the variational parameters \(\theta\).
If only single-qubit Pauli rotations are used to encode \(d\)-dimensional data \(x = (x_1, ... x_d)^T\), the spectrum is given by a symmetric, cubic \(d\)-dimensional hypergrid
\begin{equation}
\label{eq:naive_spectrum}
	\Omega = \left\{\omega \in \mathds{Z}^d \mid  -N(x_i) \leq \omega_i \leq N(x_i)\text{ } \forall i = 1,...,d \right\}.
\end{equation}
\(N(x_i)\) denotes how often the \(i-th\) component of the input vector is encoded in the circuit.

Although \cite{Schuld2021effect} states an explicit formula for the Fourier coefficients in terms of the circuit gate unitaries, it is difficult to work with as it involves computing contractions over these unitaries.
As a consequence, the functional dependence of the Fourier coefficients \(c_\omega(\theta)\) on the variational parameters \(\theta\) has been largely unexplored.
This leads to a problem however: The spectrum computed from \cref{eq:naive_spectrum} overestimates the frequencies that are available to the \gls{VQA} as some of the coefficients are constrained to be zero for all values of \(\theta\).
\Cref{fig:zero_coefs_example} represents two such constraints.
We want to stress that this is not an extreme or rare scenario, as it can be observed in typical \gls{VQC} architectures used in the literature as illustrated in Figure 5 of \cite{Schuld2021effect}.

Nemkov et al. proposed an algorithm to compute an expansion of \(f_\theta(x)\) in terms of trigonometric polynomials \cite{Nemkov_2023}, which is central to determining the exact spectrum and coefficients of a \gls{VQC}.
The method only works for \glspl{VQC} for which all constant gates are Clifford gates, the variational gates are Pauli rotations and the observable is a Pauli string (so-called Clifford + Pauli circuits).
However, we want to note that single-qubit Pauli rotations together with the CNOT gate, which is a Clifford gate, build a universal gate set.
Moreover, observables are by definition Hermitian and can therefore be decomposed into a sum of Pauli strings.
Hence, every \gls{VQC} can be transpiled into a form that is compatible with the algorithm.

In a preprocessing step, commutation relations are used to move all constant Clifford gates to the end of the circuit.
The corresponding unitaries are absorbed in the observable and the gates are removed from the circuit.

The main part of the algorithm builds on the fact that for any Pauli rotation \(R_P(\phi)\) with generator \(P\) and angle \(\phi\), the identity
\begin{equation}
R_P^\dagger(\phi) \mathcal{O} R_P(\phi) = \begin{cases}
	\mathcal{O}\text{, } & \text{if } [\mathcal{O}, P] = 0 \\
	\mathcal{O} \cos(\phi) + i P \mathcal{O} \sin(\phi) \text{, } & \text{if } [\mathcal{O}, P] \neq 0
\end{cases}
\end{equation}
holds.

This can now be used to build a binary tree as shown in \cref{algo:nemkov_tree}.
The nodes of the tree represent Clifford + Pauli circuits in their canonical normal form
\begin{equation}
    (P_1, P_2, ..., P_L | \mathcal{O}),
\end{equation}
where the \(P_j \text{, } j = 1, ..., L\) denote the generators of the Pauli rotations left in the circuit after all Clifford gates have been absorbed in the observable, which is denoted by \(\mathcal{O}\).

\begin{figure}
\centering
\includegraphics{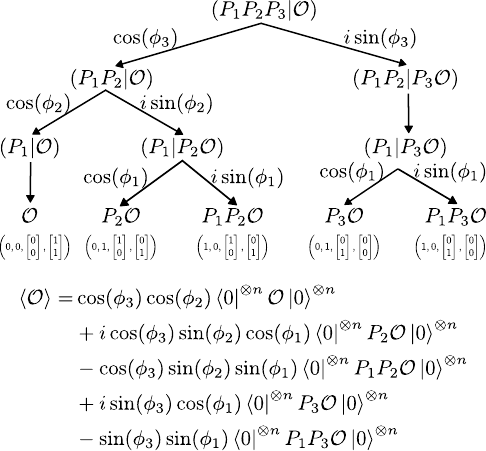}
\caption{An example computational tree generated by \cref{algo:nemkov_tree} is shown alongside the trigonometric factors that belong to each edge in the tree. In this example \([P_1, \mathcal{O}] = [P_2, P_3 \mathcal{O}] = 0\) while all other commutators are non-vanishing.
Each leaf node is labelled by a tuple \((s, c, s', c')\), where \(s, c \in \mathds{N}_0^d \text{, } s', c' \in \mathds{N}_0^w\) count the number of sine and cosine factors along the path to that node for each input parameter (\(s\) and \(c\)) and variational paramater (\(s'\) and \(c'\)) respectively.
In this example \(\phi_1\) is considered an input parameter and \(\phi_2\) and \(\phi_3\) are variational parameters.
Therefore \(d=1\) and \(w=2\).
Included below the tree is the final decomposition of the expectation value in terms of trigonometric polynomials. Each leaf node corresponds to one term with the respective coefficient being the product of the trigonometric factors along the path towards that leaf node.}
\label{fig:binary_tree_formula_example}
\end{figure}

\begin{figure}
\centering
\includegraphics[]{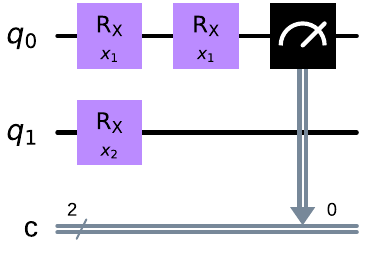}
\caption{Example circuit where certain Fourier coefficients are constrained to zero. The first feature \(x_1\) is encoded twice into the first qubit with \(R_X\) gates. The second feature \(x_2\) is only encoded into the second qubit. The \(ZI\) observable is chosen as the output. According to \cref{eq:naive_spectrum} the Fourier spectrum of this circuit is given by \(\Omega = \{-2, -1, 0, 1, 2\} \times \{-1, 0, 1\}\). However, there are two effects that eliminate frequencies from the spectrum which are ignored by \cref{eq:naive_spectrum}. Firstly, the \(x_2\)-rotation gate is outside the light cone of the measurement and can therefore not affect the outcome. And secondly, the two \(x_1\)-rotation gates combine into a single rotation around the same axis with angle \(2x_1\) \cite{Periyasamy2022}. Therefore the output of this \gls{VQC} is going to be \(\langle Z \rangle = \sin(2x_1)\), which has the Fourier spectrum \(\Omega = \{-2, 2\} \times \{0\}\).}
\label{fig:zero_coefs_example}
\end{figure}

In order to obtain the decomposition of \(f_\theta(x)\) from the binary tree one must consider every path through the tree.
Along each path the trigonometric factors on the edges are multiplied with each other.
The result of each path is multiplied again by the expectation value of the observable in the leaf node with respect to the state \(\ket{0}^{\otimes n}\).
Finally, the contributions of all paths have to be summed.
An example of this procedure is illustrated in \cref{fig:binary_tree_formula_example}.

\begin{algorithm}
\begin{algorithmic}
\STATE \textbf{Input}: \((P_1, P_2, ..., P_L | \mathcal{O})\)
\STATE Initialize empty tree
\STATE \(\text{tree.root} \gets (P_1, P_2, ..., P_L | \mathcal{O})\)
\STATE \(\text{nodesToProcess} \gets [(P_1, P_2, ..., P_L | \mathcal{O})]\)
\WHILE{nodesToProcess not empty}
    \STATE \(\text{newNodesToProcess} \gets []\)
    \FOR{node \((P_1', P_2', ..., P_{L'}' | \mathcal{O}')\) in nodesToProcess}
        \IF{\(L' > 1\)}
            \STATE Add \((P_1', P_2', ..., P_{L'-1}' | \mathcal{O}')\) to newNodesToProcess
        \ENDIF
        \STATE \(\text{node.leftChild} \gets (P_1', P_2', ..., P_{L'-1}' | \mathcal{O}')\)
        \IF{\([P_{L'}', \mathcal{O}'] \neq 0\)}
            \IF{\(L' > 1\)}
                \STATE Add \((P_1', P_2', ..., P_{L'-1}' | P_{L'}' \mathcal{O}')\) to newNodesToProcess
            \ENDIF
             \STATE \(\text{node.rightChild} \gets (P_1', P_2', ..., P_{L'-1}' | P_{L'}' \mathcal{O}')\)
             \STATE \(\text{node.leftFactor} \gets \cos(\phi_{L'})\)
             \STATE \(\text{node.rightFactor} \gets \sin(\phi_{L'})\)
         \ENDIF
    \ENDFOR
    \STATE \(\text{nodesToProcess} \gets \text{newNodesToProcess}\)
\ENDWHILE
\end{algorithmic}
\caption{Computing the binary tree from the circuit in normal form representation. At each step we process all current leaf nodes of the tree. If the last Pauli generator commutes with the observable of the node it spawns a single child. Otherwise two different children are created. If a node has no more Pauli generators, no more children are spawned. Algorithm originally proposed in \cite{Nemkov_2023}.}
\label{algo:nemkov_tree}
\end{algorithm}

\subsection{Non-uniform Fast Fourier Transform}
\label{subsec:NFFT}
Since we saw in the previous section that it is natural to describe the functions represented by \glspl{VQC} in the Fourier domain it would be beneficial to do the same with the data.
Since the function underlying the data is unknown in any practical machine learning context, we have to work with the data samples themselves to extract frequency information.
Typically one would use the \gls{FFT} algorithm \cite{cooley1965algorithm} to compute a discrete Fourier transform from the data samples.
However this is only possible for data that is evenly spaced, i.e. arranged on a regular grid.
Almost no datasets for real world applications of machine learning models have this property.

There are various algorithms that generalize the \gls{FFT} to non-uniform data \cite{fessler2003nonuniform, dutt1993fast, beylkin1995on, ware1998fast, fourmont2003nonequispaced, nieslony2003approximate}, but they deal with the inverse of our problem: sampling irregularly from a Fourier sum with regular frequencies.
As these methods are not easily inverted, one is compelled to resort to iterative numerical inversion techniques, such as those described in \cite{kunis2007stability}. The following paragraphs briefly explain how this numerical inversion is done to obtain a frequency representation of our data.

Given a regular \(d\)-dimensional lattice of frequencies
\begin{equation}
    \label{eq:infft_grid}
    I^d_N = \left\{- \frac{N_1}{2}, ..., \frac{N_1}{2} - 1\right\} \times \cdots \times  \left\{- \frac{N_d}{2}, ..., \frac{N_d}{2} - 1\right\}
\end{equation}
we search for Fourier coefficients \(f_\omega\), \(\omega \in I^d_N\) such that the resulting Fourier sum
\begin{equation}
   f(x) = \sum_{\omega \in I^d_N} f_\omega e^{2 \pi i \omega x}
\end{equation}
represents the data.
That means that for \(M\) given data samples \((x_j, y_j) \in [0, 1]^d \times \mathds{R}\) we want to achieve that \(f(x_j) \approx y_j \text{ } \forall j = 0, ..., M-1\).
In principle we could even allow for our data to be complex valued, however machine learning tasks usually deal with real quantities.

The inversion problem now boils down to solving the system of linear equations
\begin{equation}
\label{eq:linear_system}
    A f \approx y,
\end{equation}
where \(A_{j\omega} = e^{2 \pi i \omega x_j}\), \(\omega \in I^d_N\) and \(j = 0, ..., M-1\).

Going forward we will only consider the case where \cref{eq:linear_system} is under-determined as this is most often the only relevant one for our application.
In that case \cref{eq:linear_system} will have multiple suitable solutions for the coefficients \(f_\omega\).
Therefore some regularization criterion is needed to specify a unique solution.
This can be done by providing so-called damping factors \(w_\omega > 0\) for every frequency \(\omega\) and searching for the solution that minimizes a weighted \(L^2\)-norm of the Fourier coefficients
\begin{align}
    \underset{f_\omega}{\text{minimize }} & \sum_{\omega \in I^d_N} \frac{\lvert f_\omega \rvert^2}{w_\omega} \nonumber \\
    \text{subject to } & A f = y
\end{align}

The implementation provided by \cite{keiner2009using} was used for our experiments.
\section{Related Work}
\label{sec:related_work}

The Fourier representation of \glspl{VQC} has been widely used in recent years to study quantum machine learning algorithms.
For example, \cite{barthe2023gradients} prove that even though data re-uploading increases the frequency spectrum, the high-frequency coefficients will vanish exponentially.
They argue that this provides a natural protection of these models against overfitting by using this property to bound the derivative of the model output with respect to the input data.
Moreover the results in \cite{peters2023generalization} indicate that, depending on the redundancies in the spectrum, quantum models might inherit the property of \textit{benign overfitting} from classical Fourier models, where the model is both able to generalize well but also interpolate noisy training data perfectly.
\cite{mhiri2024constrained} showed that similar to the well-known phenomenon of \textit{barren plateaus} \cite{ragone2023unified}, the variance of the Fourier coefficients vanishes exponentially with increasing qubit count for some architectures by connecting the coefficient variance to the redundancy of the corresponding frequency in the spectrum.
Furthermore, \cite{jaderberg2024let} state that by including trainable weights in the encoding gates, the frequency spectrum is no longer static over the training process, but becomes learnable itself.
They argue that this facilitates \glspl{VQA} learning the appropriate frequency spectrum for the given learning task automatically with only a moderate increase in parameters.

\section{Methods}

\subsection{Calculating the exact spectrum and coefficients}
\label{sec:calculate_spectrum}
To compute the spectrum and the coefficients we make use of the algorithm from \cite{Nemkov_2023} (for more we refer to \cref{sec:background}).
The algorithm outputs a finite number of tuples \((s, c, s', c')\), with \(s, c \in \mathds{N}_0^d \text{, } s', c' \in \mathds{N}_0^w\), and constants \(k_{s, c, s', c'}\) which allow us to write the output of the \gls{VQC} in the following form:
\begin{align}
\label{eq:final_fourier_decomposition}
f_\theta(x) = &\sum_{\substack{s\text{, } c \in \mathds{N}_0^d \\
						s'\text{, } c'\in \mathds{N}_0^w}}
	k_{s, c, s', c'} 2^{-\sum_{j=1}^d (s_j + c_j)} (-i)^{\sum_{j=1}^d s_j}\times \nonumber \\
& \prod_{k = 1}^w \sin(\theta_k)^{s'_k} \cos(\theta_k)^{c'_k} \times \\
& \prod_{j = 1}^d \sum_{a = 0}^{s_j} \sum_{b=0}^{c_j} \binom{s_j}{a} \binom{c_j}{b} (-1)^{s_j - a} e^{i (2a + 2b - s_j - c_j)x_j}. \nonumber
\end{align}
The details of the derivation are contained in \cref{sec:appendix_spectrum}.
From this representation we can calculate both the spectrum \(\Omega\) and the coefficients \(c_\omega(\theta)\).
However, before we do this, we want to devote some space here to discuss the different contributions to \cref{eq:final_fourier_decomposition}.

\begin{itemize}
    \item The outermost sum runs over nodes \((s, c, s', c')\) returned by the algorithm from \cite{Nemkov_2023}. The nodes are labeled by the four non-negative integer vectors \(s, c \in \mathds{N}_0^d \text{, } s', c' \in \mathds{N}_0^w\), which correspond to the number of sine and cosine terms appearing for each input (\(s \text{ and } c\)) and variational parameter (\(s' \text{ and } c'\)).
    \item The first line of \cref{eq:final_fourier_decomposition} contains some node-specific constant (w.r.t. the model parameters and inputs) prefactors. 
    The \(k_{s, c, s', c'}\) are specific \(\ket{0}^{\otimes n}\)-expectation values (confer \cref{sec:background} for more details), which are either zero or have an absolute value of one. Therefore the prefactors are dominated by the exponential decay in the L1-norm of \(s\) and \(c\).
    \item The second line consists of trigonometric polynomials of the variational parameters \(\theta\). This is the only dependence of \(f_\theta(x)\) on \(\theta\). Since this part is unique for each term of the outermost sum with different \(s'\) or \(c'\), these can never cancel each other for all values of \(\theta\).
    \item The third line is in of itself again a Fourier sum with combinatorial coefficients. It can be read off that the spectrum \(\Omega(s, c)\) of the contribution of each single node needs to be contained in the set \(\{\omega \in \mathds{N}^d \mid \omega + s + c \in \left(2 \mathds{N}\right)^d \text{, } \lvert \omega_j \rvert \leq s_j + c_j\}\). However, this does not take into account that different parts of the Fourier sum with the same exponential factor could cancel each other.
\end{itemize}

In order to find out which frequencies survive inside the inner Fourier sum of each node, we calculate the sum of all combinatorial coefficients belonging to a specific frequency \(\omega\):
\begin{equation}
    \label{eq:combinatorial_coefficients_sum}
    p(s, c, \omega) = \prod_{j = 1}^d \sum_{a = 0}^{s_j} \sum_{b=0}^{c_j} \binom{s_j}{a} \binom{c_j}{b} (-1)^{s_j - a} \delta^{\omega_j}_{2a + 2b - s_j - c_j}.
\end{equation}
If this expression reduces to zero, the frequency does not belong to the spectrum of the Fourier sum of this node.
That is obviously the case whenever for any index \(j\) \(\omega_j \notin [-c_j -s_j, c_j + s_j]\) or \(s_j + c_j + \omega_j\notin 2\mathds{Z}\).
In all other cases, \cref{eq:combinatorial_coefficients_sum} can be evaluated in terms of hypergeometric functions (see \cref{sec:appendix_spectrum}).

Therefore we can finally write down the frequencies that a given node contributes to the spectrum
\begin{align}
    \Omega(s, c) = \{\omega \in \mathds{N}^d \mid & \omega + s + c \in \left(2 \mathds{N}\right)^d \text{, } \lvert \omega_j \rvert \leq s_j + c_j \text{, } \nonumber \\
    & p(s, c, \omega) \neq 0\}.
\end{align}

Since the final output of the \gls{VQC} is itself a weighted sum of each of the Fourier sums belonging to the computational nodes (cf. \cref{eq:final_fourier_decomposition}), we need to take into account that for some frequencies the contributions of the different nodes might cancel each other.
As mentioned above, this is only possible for nodes that share the same \(s'\) and \(c'\).

The overall spectrum \(\Omega\) of the quantum model is just the union of the spectra of each contributing node, for which the contributions of all nodes do not cancel exactly
\begin{align}
    \label{eq:final_spectrum}
    & \Omega = \left\{ \left. \omega \in \bigcup_{s, c \in \mathds{N}_0^d} \Omega (s, c) \right\vert \exists s', c' \text{ s.t. } \right. \\
     & \left. \sum_{s, c \in \mathds{N}_0^d}k_{s, c, s', c'} 2^{-\sum_{j=1}^d (s_j + c_j)} (-i)^{\sum_{j=1}^d s_j} p(s, c, \omega) \neq 0 \right\}. \nonumber
\end{align}

\Cref{eq:combinatorial_coefficients_sum} also enables us to explicitly write down the Fourier coefficient belonging to each frequency
\begin{align}
\label{eq:fourier_coefficients}
c_\omega(\theta) = \sum_{\substack{s\text{, } c \in \mathds{N}_0^d \\
						s'\text{, } c'\in \mathds{N}_0^w}}
	& k_{s, c, s', c'} 2^{-\sum_{j=1}^d (s_j + c_j)} (-i)^{\sum_{j=1}^d s_j} \times \nonumber \\[-1em]
 & p(s, c, \omega) \prod_{k = 1}^w \sin(\theta_k)^{s'_k} \cos(\theta_k)^{c'_k}.
\end{align}


To summarize the most important points: The function that a specific \gls{VQC} realizes can be decomposed as a Fourier sum of the input parameters. The spectrum of this Fourier sum is given by \cref{eq:final_spectrum} and the corresponding coefficients for each frequency can be calculated using \cref{eq:fourier_coefficients}.

\subsection{Ranking VQC architectures for a given dataset}
\label{sec:ranking}
To motivate the experiments we want to establish the following two ideas:
Firstly, in order to have the ability to represent the data well a \gls{VQC} should include the \textit{dominant} frequencies of the dataset in its spectrum.
For example, sharp jumps in the data typically require high-frequency components in the direction of the jump.
Secondly, there is a trade-off between architecture expressivity and trainability \cite{holmes2022connecting}.
Since larger Fourier spectra typically come with higher expressivity \cite{Schuld2021effect} we expect that architectures with broader spectra will be harder to train.
Therefore an optimal choice of architecture should include as many (or as large) frequencies as necessary to accurately describe the data but at the same time as few as possible to remain trainable.
In order to determine the best architecture for these datasets the following steps are taken:

First use \cref{algo:nemkov_tree} and \cref{eq:final_spectrum} to determine the frequency spectrum of every \gls{VQC} architecture.

Secondly, use the inverse \gls{NFFT} described in \cref{subsec:NFFT} to obtain a frequency-space representation of the data.
In order to do this, a grid in the frequency domain has to be chosen beforehand. 
The grid defined by \cref{eq:naive_spectrum} presents a natural choice at this step.
However, one needs to take care since the inverse \gls{NFFT} by definition works with asymmetric grids (see \cref{eq:infft_grid}) while the grids defined by \cref{eq:naive_spectrum} are always symmetric around the origin.
As a result an additional gridpoint along each dimension is needed.
Further, one needs to deal with the non-uniqueness of the inverse \gls{NFFT} if the grid is larger than the dataset.
In this case, damping factors \(w_k\) need to be introduced as explained in \cref{subsec:NFFT}.
Since we are interested in how well a \gls{VQC} with a given spectrum \(\Omega\) is able to represent the data, we are looking for a spectral representation of the data that tries to use as few frequencies outside of \(\Omega\) as possible.
Therefore we choose the damping factors as
\begin{equation}
    w_\omega = \begin{cases}
        10^3 \text{, } & \omega \in \Omega \\
        10^{-3} \text{, } & \omega \notin \Omega.
    \end{cases}
\end{equation}

At this point there can be multiple indications that the \gls{VQC} is not well suited for the dataset.
\begin{enumerate}
    \item The grid provided by \cref{eq:naive_spectrum} is too small and the inverse NFFT therefore converges to an inaccurate frequency representation. This can be quantified by the residual
    \begin{equation}
        R_{\text{NFFT}} = \sum_{j = 0}^{M-1} \left\lVert y_j - \sum_{\omega \in I_N^d} f_\omega e^{2\pi i \omega x} \right\rVert^2.
    \end{equation}
    A large \(R_{\text{NFFT}}\) indicates that the input encoding needs to be repeated more often, as even arbitrary access to all possible frequencies does not yield a good representation of the data.
    However, we will only be comparing circuits with the same amount of input encoding in this work, therefore we will not consider \(R_{\text{NFFT}}\) further.
    \item The dataset has large Fourier coefficients on frequencies which are not supported by the \gls{VQC} architecture.
    This is quantified by
    \begin{equation}
        R_{\Omega} = \sum_{\omega \in I_N^d \setminus \Omega} \lvert f_\omega \rvert^2. 
    \end{equation}
    \item No choice of the variational parameters reproduces the pattern of the coefficients \(f_\omega\) through \cref{eq:fourier_coefficients}. However determining the best possible fit
    \begin{equation}
        \label{eq:coef_mismatch_exact}
        \underset{\theta}{\min} \sum_{\omega \in \Omega} \lvert f_\omega - c_\omega(\theta) \rvert^2.
    \end{equation}
    is infeasible, since it is equivalent to training the \gls{VQC} model.
    Therefore a different heuristic is needed, which captures the same information.
    We suspect that the two main reasons for which a specific combination of coefficients cannot be reached by the \gls{VQC} is either that some coefficients lie outside the range of the corresponding trigonometric polynomials or correlations between different polynomials prohibit that specific combination of values.
    An easy example to illustrate the latter point is \(c_{\omega_1}(\theta) = \cos(\theta)\) and \(c_{\omega_2}(\theta) = -\cos(\theta)\), for which there exists no parameter \(\theta\) such that \(c_{\omega_1}(\theta) = c_{\omega_2}(\theta) = 1\).
\end{enumerate}

In order to get an estimate on the effect of issue 3) on the task at hand, we need to estimate the correlation between coefficients.
Given the matrix \(\Sigma_{\omega_1, \omega_2}\) of covariances and the vector \(\bar{c}_\omega\) of means, which define the best multivariate gaussian fit to the true distribution of the coefficients
the Mahalanobis distance \cite{mahalanobis1936generalised}
\begin{equation}
    R_\text{corr} = \sqrt{(f - \bar{c})^\dagger \Sigma^{-1}(f - \bar{c})}
\end{equation}
provides a measure for the chance of observing a specific set of values \(f_\omega\) for the coefficients \(c_\omega(\theta)\) from this best fit gaussian distribution.
\Cref{sec:appendix_covariance} details how \(\Sigma_{\omega_1, \omega_2}\) and \(\bar{c}_\omega\) are calculated.

We want to stress that this is only a heuristic, as the true distribution underlying \(c_\omega(\theta)\) is a lot more complex.
To illustrate this point further, consider the simple, but not unrealistic example of \(c_{\omega_1}(\theta) = a\sin(\theta)\) and \(c_{\omega_2}(\theta) = a\cos(\theta)\).
As is easily seen from \cref{eq:powers_of_trigs_integral}, the off-diagonals of the covariance matrix and the means vanish.
Therefore the values \(f_{\omega_1} = 0\) and \(f_{\omega_2} = 0\) yield \(R_\text{corr} = 0\), which should indicate a high chance of being able to achieve these values.
However, the true distribution of the \(c_\omega(\theta)\) is a circle with radius \(a\) around \((0, 0)\).
Therefore the coefficients will always be arbitrarily far away from the desired target.
We discuss how well this heuristic performs in practice in \cref{sec:results}.

Finally, we also introduce a punishment term for excessively large spectra
\begin{equation}
    R_\text{punish} = \frac{\lvert \Omega \rvert}{\lvert I^d_N \rvert},
\end{equation}
due to the expressivity-trainability tradeoff.

We give every architecture a score, which is calculated by normalizing the values for \(R_\Omega\) and \(R_\text{corr}\) and adding them with \(R_\text{punish}\), which is already normalized by definition.
Ultimately, we rank the architectures according to their score.
The way the score has been defined means that circuits with a lower score are predicted to fit the data better.
Therefore we assign the best ranking to the lowest score.

\subsection{Experimental Setup}
\label{sec:experimental_setup}

\begin{figure*}
    \centering
    \includegraphics[scale=0.82]{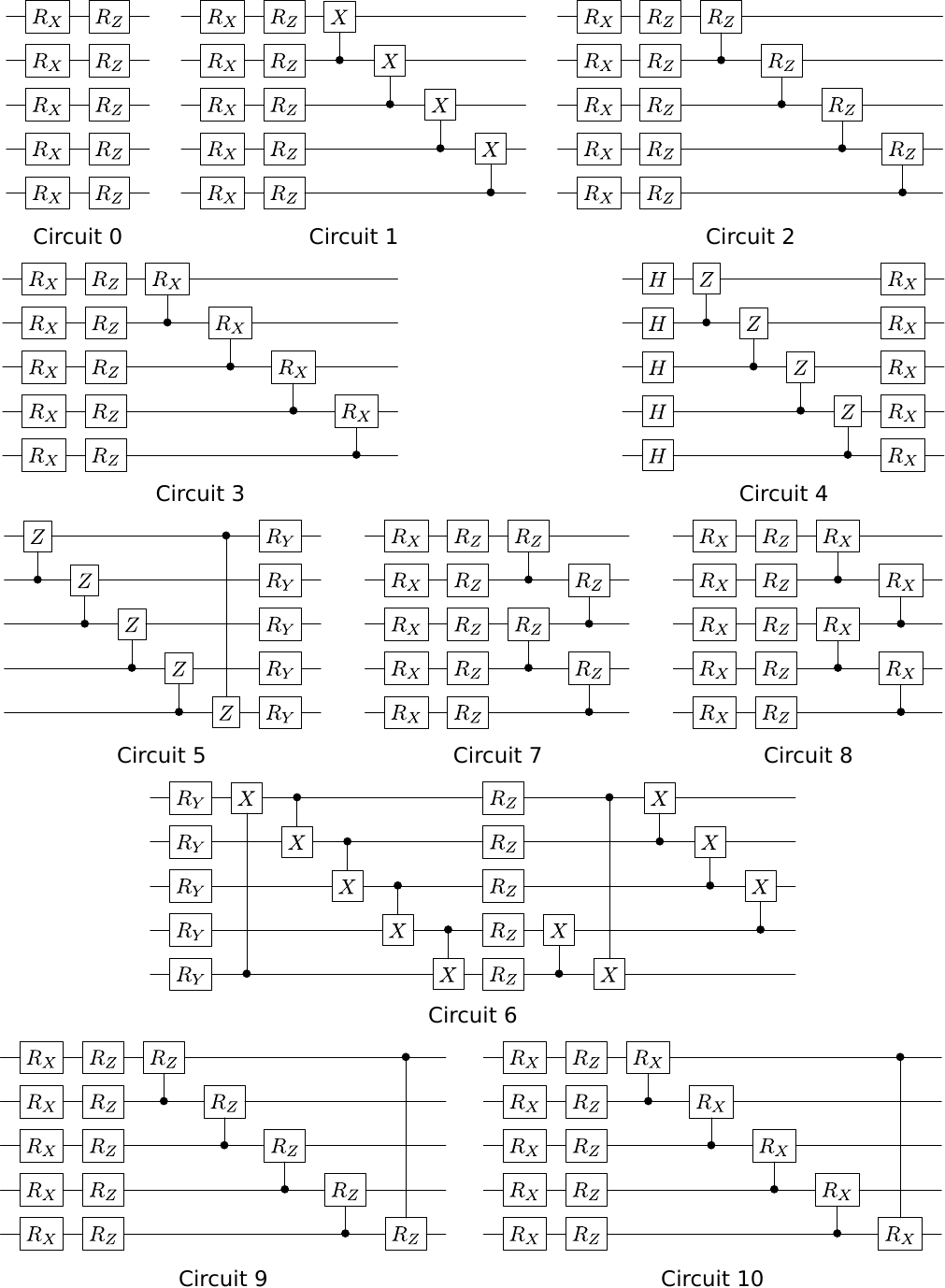}
    \caption{Entangling layers used for the circuits in the experiments. In each circuit, the entangling layer is repeated three times with single-qubit rotation gates as encoding layers in between. The type of rotation was chosen such that no rotation gate is repeated twice in a row. The output of the circuit is given by measuring the \(Z_3\) observable. Depicted are the architectures used for the 5-dimensional Friedman dataset. For the MNIST dataset, 4 qubit equivalents to the depicted circuits were used and each of the two features was encoded on two qubits per layer. The choice of circuits was inspired by \cite{sim2019expressibility}.}
    \label{fig:circuits}
\end{figure*}

Now that we have established how a full characterization of the Fourier sum that any given \gls{VQC} represents can be computed, we want to test whether this information can be used to predict which architecture is best suited for a given dataset.

Before we go into the details of the experiments that were conducted, we want to note that once \cref{algo:nemkov_tree} has been applied to any given circuit it is possible to classically simulate that circuit in a time polynomial in the number of leaf nodes of the arising tree.
Therefore it is clear that for generic (non-Clifford) circuits the tree will be exponentially large.
Consequently obtaining the full information on the Fourier spectrum and the coefficients will also require exponential effort.
For this reason the experiments are going to remain in the realm of small circuits with 4 qubits and no more than three layers.
The experiments should the viewed as a proof of concept: while the methods we present here are certainly not directly scalable to larger circuits it is conceivable that suitable approximations to our methods can be found more efficiently.
However investigating this potential is outside the scope of this work and will be left for future study.

We want to demonstrate that it is possible to exploit the knowledge of the \gls{VQC}s Fourier decomposition and dataset frequencies to predict the training performance.
For this purpose we consider a regression problem introduced by Friedman \cite{friedman1991multivariate} with the ground-truth function given by
\begin{equation}
    y(x) = 10 \sin(\pi x_1 x_2) + 20 (x_3 - \frac{1}{2})^2 + 10 x_4 + 5 x_5
\end{equation}
and the MNIST handwritten digits dataset \cite{fisher1936use} as a regression task where the labels of the different classes are mapped linearly into the interval of \([0, 1]\).
Since the dimension of the MNIST dataset is too large to encode the data into a \gls{VQC} directly, we instead perform dimensionality reduction via t-SNE \cite{van2008visualizing} and map the dataset into a 2D space beforehand.

We train a set of eleven different architectures on these datasets.
The quantum circuits consist of one of the variational layers depicted in \cref{fig:circuits}, which is repeated three times.
The expressivity and entangling capability of these varaitional layers has been previously studied in \cite{sim2019expressibility}.
Each entangling layer is preceded by single-qubit Pauli \(X\) or \(Y\) rotation gates which encode the classical input data.
The encoding rotation axis is always chosen to be different from the first rotation gates of the variational layer to avoid successive rotations around the same axis.
Finally, the expectation value of the resulting quantum state with respect to the \(Z_3\) observable constitutes the output of the model.

Next, we train these \glspl{VQC} on the above mentioned datasets with 5000 samples for 100 epochs with a batch size of 128 using the Adam optimizer \cite{kingma2014adam} with a learning rate of \(0.005\) and \gls{MSE} loss function. Their performances are continuously validated on 500 samples of unseen test data during the training.
The gradients of the \gls{VQC} are evaluated using the parameter shift rule \cite{Schuld19evaluating}.
As we are not interested on the absolute performance of each architecture but rather how they perform in relation to each other, we abstain from hyperparameter tuning.

Finally, we rank the set of \glspl{VQC} for these datasets according to \cref{sec:ranking}.
However, the Mahalanobis distance is only evaluated for a randomly selected subset of 100 coefficients, since calculating the full covariance matrix is too expensive.

The results of these experiments are discussed in the following section.

\section{Results and Discussion}
\label{sec:results}

\subsection{Effect of the architecture on the VQC spectrum}

\begin{figure}
    \centering
    \includegraphics{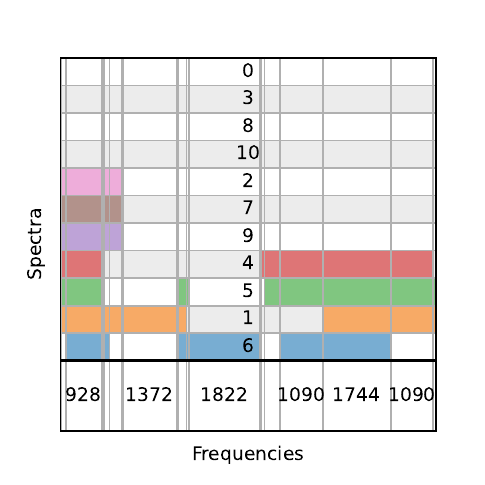}
    \caption{Visualization of the VQC spectra. Each row represents one of the spectra of the architectures zero to ten from \cref{fig:circuits}.
    The x-axis consists of a number of cells. Each cell represents the intersection of a unique combination of spectra. These spectra appear as colored in the column belonging to the cell. The number of frequencies contained in the intersection is indicated by the width of the respective column and explicitly given by the number in the bottommost cell.
    The numbers are only shown in the larger cells for readability.
    For example, in the column with the label 1822 only the blue row is colored. This means that the spectrum of architecture six contains 1822 frequencies, which are not present in any other spectrum.
    The rightmost column with the label 1090 indicates that there are 1090 frequencies unique to the architectures one, four and five and so on. 
    The spectra belonging to architectures zero, three, eight and ten only contain seven frequencies and therefore appear to be empty on this scale.
    We want to emphasize that all architectures have the same number of encoding gates and measurement observable.
    The differences in the spectra stem purely from the different variational layers.}
    \label{fig:spectra_venn}
\end{figure}

The main point of this work is to demonstrate that the Fourier spectrum of a \gls{VQC} does not only depend on the encoding, but rather on the entire circuit.
This fact is illustrated by \cref{fig:spectra_venn}, which shows an intersection diagram of the spectrum of each of the circuits depicted in \cref{fig:circuits}.
The spectra of circuits zero, three, eight and ten only consist of the same seven frequencies, namely
\begin{align*}
    \Omega = \{ & (0, 0, -3, 0, 0), (0, 0, -2, 0, 0), \\
    & (0, 0, -1, 0, 0), (0, 0, 0, 0, 0), \\
    & (0, 0, 1, 0, 0), (0, 0, 2, 0, 0), (0, 0, 3, 0, 0) \}.
\end{align*}
Circuits two, seven and nine and circuits four and five have very similar spectra respectively.
However, the latter are almost four times as broad as the former.
Circuits six and one admit the largest spectra, with over 1000 frequencies being unique to each.

Since in all cases the encoding consisted of single-qubit Pauli rotations, which were repeated thrice throughout the circuit, it would be expected that all circuits share the same spectrum \(\Omega = \{-3, -2, -1, 0, 1, 2, 3\}^{\times 5}\).
However, as can be seen from \cref{fig:spectra_venn}, the different variational parts of the circuit each constrain different sets of coefficients to zero, therefore removing the corresponding frequency from the spectrum.
In some cases the reason for these constraints is obvious, e.g. circuit 0 does not contain any entangling gates.
Therefore most of the circuit lies outside the light cone of the local \(Z_3\) observable and therefore cannot contribute to the spectrum (also cf. \cref{fig:zero_coefs_example}).
However, the missing frequencies cannot be explained by the heuristics depicted in \cref{fig:zero_coefs_example} for any other cases.

The sizes of the spectra range from seven up to 6125 frequencies. 
Furthermore there is no trivial order of inclusion among them.
This fact underpins the point that even for practically relevant circuits, the spectrum cannot be determined by the encoding alone.

\subsection{Predicting model performance}

\begin{figure*}
\centering
\begin{subfigure}{.49\textwidth}
    \centering
    \includegraphics{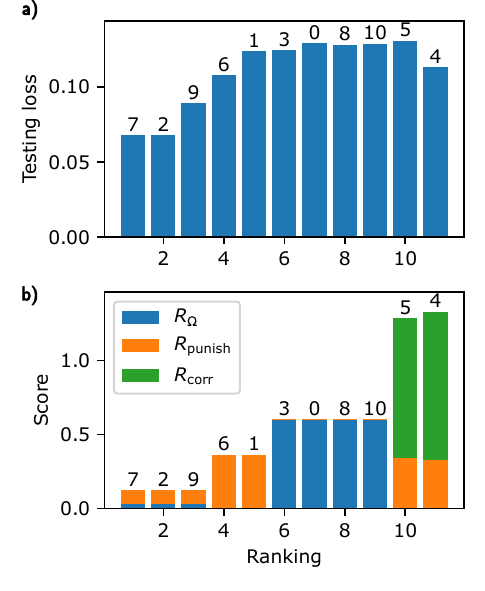}
\end{subfigure}
\begin{subfigure}{.49\textwidth}
    \centering
    \includegraphics{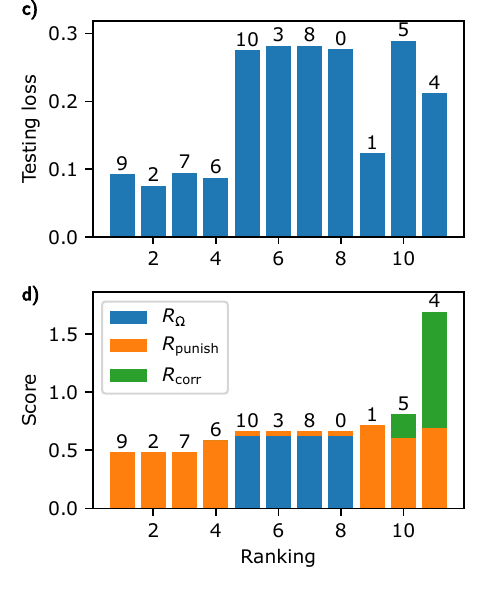}
\end{subfigure}
\caption{Ranking of the architectures compared with performance on the test data and composition of the score given to each circuit. The figure shows results on the Friedman datasets in a) and b) and on the MNIST dataset in c) and d). The top plots a) and c) depict the minimally achieved testing loss of each circuit. The bottom plots b) and d) show the score that was assigned to each architecture. Its composition in terms of \(R_\Omega\), \(R_\text{punish}\) and \(R_\text{corr}\) is indicated by color. We want to emphasize that a lower score means we predict the model to fit the data better. The number above each bar represents the label of the architecture from \cref{fig:circuits}. The x-axis corresponds to the spot on the ranking that this architecture achieved.}
\label{fig:ranking_chart}
\end{figure*}

Finally, we test the ranking method described in \cref{sec:ranking}.
\Cref{fig:ranking_chart} depicts the minimal loss each model achieved on the test data as the height of the corresponding bar as well as the size of each of the three contributions \(R_\Omega\), \(R_\text{punish}\) and \(R_\text{corr}\) towards the final score.
The bars are arranged according to the ranking computed for these architectures.
The upper pair of figures shows the results from the Friedman dataset, the lower one from the MNIST dataset.

In both experiments the architectures two, seven and nine received the best, almost identical scores.
On the Friedman dataset they also achieved the best testing losses and in the MNIST case they are among the top four, with architecture six only getting scored slightly worse but achieving a similar performance.
Circuits zero, three, eight and ten were the ones with the smallest spectra, only containing the same seven frequencies each.
Therefore they get by far the highest contributions to their scores from \(R_\Omega\).
Interestingly, these are not the worst performing circuits, as architecture five only achieved a similarly bad test loss on both datasets.
This is somewhat surprising, as architecture five has the third largest spectrum.
However, the high contribution of \(R_\text{corr}\) to its score indicates that there is a large amount of correlation between the coefficients.
Circuit four also has a large contribution of \(R_\text{corr}\) to it's score and is ranked worse than circuit five in both cases.
However it does indeed perform better on both datasets.
This might indicate that the current estimate of \(R_\text{corr}\) from 30 coefficients is not yet accurate enough to properly distinguish these two architectures.

Furthermore, the ratings in the MNIST case are almost all very close to each other.
For example, architecture one only received a slightly worse ranking than zero, three, eight and ten, while performing significantly better.
It seems like that \(R_\Omega\) and \(R_\text{corr}\) are more important factors to the final performance than \(R_\text{punish}\).
This leads to the question of how these factors should be weighted to give an even more accurate picture.
However this lies outside the scope of this paper and is left for future research.

\section{Conclusion and outlook}
\label{sec:conclusion}
In this paper, we showed that the variational part of a \gls{VQC} does indeed significantly affect its Fourier spectrum. The spectrum is therefore not purely determined by its encoding gates but rather by the whole circuit.

Building upon previous work from Nemkov et al. \cite{Nemkov_2023} we presented an algorithm to compute the exact frequency spectrum available to any given \gls{VQC}. Furthermore we showed that the Fourier coefficients are given by trigonometric polynomials of the variational parameters and provided an explicit formula to compute any given coefficient.

We applied our methods on a set of architectures, which are commonly used in the quantum machine learning community.
There we underpinned the importance of the impact of the variational part of the circuit for the spectrum, as the architectures yielded widely different spectra even though they use the same encoding.

Finally we demonstrated that the complete knowledge of the spectrum can be exploited to select the best choice of architecture for a given dataset.

However, the presented method for calculating the Fourier spectrum of a \gls{VQC} is at least as computationally hard as classically simulating the same circuit.
Furthermore, we only investigated how different variational layers impact the spectrum.
In principle the same study could be repeated with the choice of measurement observable, as it too influences the available spectrum.

\bibliographystyle{IEEEtran}
\bibliography{references}

\begin{thebibliography}{10}
\providecommand{\url}[1]{#1}
\csname url@samestyle\endcsname
\providecommand{\newblock}{\relax}
\providecommand{\bibinfo}[2]{#2}
\providecommand{\BIBentrySTDinterwordspacing}{\spaceskip=0pt\relax}
\providecommand{\BIBentryALTinterwordstretchfactor}{4}
\providecommand{\BIBentryALTinterwordspacing}{\spaceskip=\fontdimen2\font plus
\BIBentryALTinterwordstretchfactor\fontdimen3\font minus \fontdimen4\font\relax}
\providecommand{\BIBforeignlanguage}[2]{{%
\expandafter\ifx\csname l@#1\endcsname\relax
\typeout{** WARNING: IEEEtran.bst: No hyphenation pattern has been}%
\typeout{** loaded for the language `#1'. Using the pattern for}%
\typeout{** the default language instead.}%
\else
\language=\csname l@#1\endcsname
\fi
#2}}
\providecommand{\BIBdecl}{\relax}
\BIBdecl

\bibitem{peruzzo2014variational}
A.~Peruzzo, J.~McClean, P.~Shadbolt, M.-H. Yung, X.-Q. Zhou, P.~J. Love, A.~Aspuru-Guzik, and J.~L. O’brien, ``A variational eigenvalue solver on a photonic quantum processor,'' \emph{Nature communications}, vol.~5, no.~1, p. 4213, 2014.

\bibitem{kandala2017hardware}
A.~Kandala, A.~Mezzacapo, K.~Temme, M.~Takita, M.~Brink, J.~M. Chow, and J.~M. Gambetta, ``Hardware-efficient variational quantum eigensolver for small molecules and quantum magnets,'' \emph{nature}, vol. 549, no. 7671, pp. 242--246, 2017.

\bibitem{parrish2019quantum}
R.~M. Parrish, E.~G. Hohenstein, P.~L. McMahon, and T.~J. Mart{\'\i}nez, ``Quantum computation of electronic transitions using a variational quantum eigensolver,'' \emph{Physical review letters}, vol. 122, no.~23, p. 230401, 2019.

\bibitem{li2019variational}
Y.~Li, J.~Hu, X.-M. Zhang, Z.~Song, and M.-H. Yung, ``Variational quantum simulation for quantum chemistry,'' \emph{Advanced Theory and Simulations}, vol.~2, no.~4, p. 1800182, 2019.

\bibitem{zhang2021shallow}
F.~Zhang, N.~Gomes, N.~F. Berthusen, P.~P. Orth, C.-Z. Wang, K.-M. Ho, and Y.-X. Yao, ``Shallow-circuit variational quantum eigensolver based on symmetry-inspired hilbert space partitioning for quantum chemical calculations,'' \emph{Physical Review Research}, vol.~3, no.~1, p. 013039, 2021.

\bibitem{schuld2015introduction}
M.~Schuld, I.~Sinayskiy, and F.~Petruccione, ``An introduction to quantum machine learning,'' \emph{Contemporary Physics}, vol.~56, no.~2, pp. 172--185, 2015.

\bibitem{biamonte2017quantum}
J.~Biamonte, P.~Wittek, N.~Pancotti, P.~Rebentrost, N.~Wiebe, and S.~Lloyd, ``Quantum machine learning,'' \emph{Nature}, vol. 549, no. 7671, pp. 195--202, 2017.

\bibitem{schuld2020circuit}
M.~Schuld, A.~Bocharov, K.~M. Svore, and N.~Wiebe, ``Circuit-centric quantum classifiers,'' \emph{Physical Review A}, vol. 101, no.~3, p. 032308, 2020.

\bibitem{guan2021quantum}
W.~Guan, G.~Perdue, A.~Pesah, M.~Schuld, K.~Terashi, S.~Vallecorsa, and J.-R. Vlimant, ``Quantum machine learning in high energy physics,'' \emph{Machine Learning: Science and Technology}, vol.~2, no.~1, p. 011003, 2021.

\bibitem{meyer2023quantum}
N.~Meyer, D.~Scherer, A.~Plinge, C.~Mutschler, and M.~Hartmann, ``Quantum policy gradient algorithm with optimized action decoding,'' in \emph{International Conference on Machine Learning}.\hskip 1em plus 0.5em minus 0.4em\relax PMLR, 2023, pp. 24\,592--24\,613.

\bibitem{wu2020quantum}
S.~Wu, S.~Jin, D.~Wen, D.~Han, and X.~Wang, ``Quantum reinforcement learning in continuous action space,'' \emph{arXiv preprint arXiv:2012.10711}, 2020.

\bibitem{jerbi2021parametrized}
S.~Jerbi, C.~Gyurik, S.~Marshall, H.~Briegel, and V.~Dunjko, ``Parametrized quantum policies for reinforcement learning,'' \emph{Advances in Neural Information Processing Systems}, vol.~34, pp. 28\,362--28\,375, 2021.

\bibitem{meyer2022survey}
N.~Meyer, C.~Ufrecht, M.~Periyasamy, D.~D. Scherer, A.~Plinge, and C.~Mutschler, ``A survey on quantum reinforcement learning,'' \emph{arXiv preprint arXiv:2211.03464}, 2022.

\bibitem{cherrat2023quantum}
E.~A. Cherrat, I.~Kerenidis, and A.~Prakash, ``Quantum reinforcement learning via policy iteration,'' \emph{Quantum Machine Intelligence}, vol.~5, no.~2, p.~30, 2023.

\bibitem{periyasamy2023batch}
M.~Periyasamy, M.~H{\"o}lle, M.~Wiedmann, D.~D. Scherer, A.~Plinge, and C.~Mutschler, ``Batch quantum reinforcement learning,'' \emph{arXiv preprint arXiv:2305.00905}, 2023.

\bibitem{Schuld2021effect}
M.~Schuld, R.~Sweke, and J.~J. Meyer, ``Effect of data encoding on the expressive power of variational quantum-machine-learning models,'' \emph{Physical Review A}, vol. 103, no.~3, p. 032430, 2021.

\bibitem{holmes2022connecting}
Z.~Holmes, K.~Sharma, M.~Cerezo, and P.~J. Coles, ``Connecting ansatz expressibility to gradient magnitudes and barren plateaus,'' \emph{PRX Quantum}, vol.~3, no.~1, p. 010313, 2022.

\bibitem{Nemkov_2023}
\BIBentryALTinterwordspacing
N.~A. Nemkov, E.~O. Kiktenko, and A.~K. Fedorov, ``Fourier expansion in variational quantum algorithms,'' \emph{Physical Review A}, vol. 108, no.~3, Sep. 2023. [Online]. Available: \url{http://dx.doi.org/10.1103/PhysRevA.108.032406}
\BIBentrySTDinterwordspacing

\bibitem{Periyasamy2022}
\BIBentryALTinterwordspacing
M.~Periyasamy, N.~Meyer, C.~Ufrecht, D.~D. Scherer, A.~Plinge, and C.~Mutschler, ``Incremental data-uploading for full-quantum classification,'' in \emph{2022 IEEE International Conference on Quantum Computing and Engineering (QCE)}.\hskip 1em plus 0.5em minus 0.4em\relax Los Alamitos, CA, USA: IEEE Computer Society, sep 2022, pp. 31--37. [Online]. Available: \url{https://doi.ieeecomputersociety.org/10.1109/QCE53715.2022.00021}
\BIBentrySTDinterwordspacing

\bibitem{cooley1965algorithm}
J.~W. Cooley and J.~W. Tukey, ``An algorithm for the machine calculation of complex fourier series,'' \emph{Mathematics of computation}, vol.~19, no.~90, pp. 297--301, 1965.

\bibitem{fessler2003nonuniform}
J.~Fessler and B.~Sutton, ``Nonuniform fast fourier transforms using min-max interpolation,'' \emph{IEEE Transactions on Signal Processing}, vol.~51, no.~2, pp. 560--574, 2003.

\bibitem{dutt1993fast}
\BIBentryALTinterwordspacing
A.~Dutt and V.~Rokhlin, ``Fast fourier transforms for nonequispaced data,'' \emph{SIAM Journal on Scientific Computing}, vol.~14, no.~6, pp. 1368--1393, 1993. [Online]. Available: \url{https://doi.org/10.1137/0914081}
\BIBentrySTDinterwordspacing

\bibitem{beylkin1995on}
\BIBentryALTinterwordspacing
G.~Beylkin, ``On the fast fourier transform of functions with singularities,'' \emph{Applied and Computational Harmonic Analysis}, vol.~2, no.~4, pp. 363--381, 1995. [Online]. Available: \url{https://www.sciencedirect.com/science/article/pii/S1063520385710263}
\BIBentrySTDinterwordspacing

\bibitem{ware1998fast}
\BIBentryALTinterwordspacing
A.~F. Ware, ``Fast approximate fourier transforms for irregularly spaced data,'' \emph{SIAM Review}, vol.~40, no.~4, pp. 838--856, 1998. [Online]. Available: \url{https://doi.org/10.1137/S003614459731533X}
\BIBentrySTDinterwordspacing

\bibitem{fourmont2003nonequispaced}
\BIBentryALTinterwordspacing
K.~Fourmont, ``Non-equispaced fast fouriertransforms with applicationsto tomography,'' \emph{Journal of Fourier Analysis and Applications}, vol.~9, no.~5, pp. 431--450, Sep 2003. [Online]. Available: \url{https://doi.org/10.1007/s00041-003-0021-1}
\BIBentrySTDinterwordspacing

\bibitem{nieslony2003approximate}
\BIBentryALTinterwordspacing
A.~Nieslony and G.~Steidl, ``Approximate factorizations of fourier matrices with nonequispaced knots,'' \emph{Linear Algebra and its Applications}, vol. 366, pp. 337--351, 2003, special issue on Structured Matrices: Analysis, Algorithms and Applications. [Online]. Available: \url{https://www.sciencedirect.com/science/article/pii/S0024379502004962}
\BIBentrySTDinterwordspacing

\bibitem{kunis2007stability}
\BIBentryALTinterwordspacing
S.~Kunis and D.~Potts, ``Stability results for scattered data interpolation by trigonometric polynomials,'' \emph{SIAM Journal on Scientific Computing}, vol.~29, no.~4, pp. 1403--1419, 2007. [Online]. Available: \url{https://doi.org/10.1137/060665075}
\BIBentrySTDinterwordspacing

\bibitem{keiner2009using}
J.~Keiner, S.~Kunis, and D.~Potts, ``Using nfft 3---a software library for various nonequispaced fast fourier transforms,'' \emph{ACM Transactions on Mathematical Software (TOMS)}, vol.~36, no.~4, pp. 1--30, 2009.

\bibitem{barthe2023gradients}
A.~Barthe and A.~P{\'e}rez-Salinas, ``Gradients and frequency profiles of quantum re-uploading models,'' \emph{arXiv preprint arXiv:2311.10822}, 2023.

\bibitem{peters2023generalization}
E.~Peters and M.~Schuld, ``Generalization despite overfitting in quantum machine learning models,'' \emph{Quantum}, vol.~7, p. 1210, 2023.

\bibitem{mhiri2024constrained}
H.~Mhiri, L.~Monbroussou, M.~Herrero-Gonzalez, S.~Thabet, E.~Kashefi, and J.~Landman, ``Constrained and vanishing expressivity of quantum fourier models,'' \emph{arXiv preprint arXiv:2403.09417}, 2024.

\bibitem{ragone2023unified}
M.~Ragone, B.~N. Bakalov, F.~Sauvage, A.~F. Kemper, C.~O. Marrero, M.~Larocca, and M.~Cerezo, ``A unified theory of barren plateaus for deep parametrized quantum circuits,'' \emph{arXiv preprint arXiv:2309.09342}, 2023.

\bibitem{jaderberg2024let}
B.~Jaderberg, A.~A. Gentile, Y.~A. Berrada, E.~Shishenina, and V.~E. Elfving, ``Let quantum neural networks choose their own frequencies,'' \emph{Physical Review A}, vol. 109, no.~4, p. 042421, 2024.

\bibitem{mahalanobis1936generalised}
\BIBentryALTinterwordspacing
P.~Mahalanobis, ``Reprint of: Mahalanobis, p.c. (1936) "on the generalised distance in statistics.",'' \emph{Sankhya A}, vol.~80, no.~1, pp. 1--7, Dec 2018. [Online]. Available: \url{https://doi.org/10.1007/s13171-019-00164-5}
\BIBentrySTDinterwordspacing

\bibitem{sim2019expressibility}
S.~Sim, P.~D. Johnson, and A.~Aspuru-Guzik, ``Expressibility and entangling capability of parameterized quantum circuits for hybrid quantum-classical algorithms,'' \emph{Advanced Quantum Technologies}, vol.~2, no.~12, p. 1900070, 2019.

\bibitem{friedman1991multivariate}
J.~H. Friedman, ``Multivariate adaptive regression splines,'' \emph{The annals of statistics}, vol.~19, no.~1, pp. 1--67, 1991.

\bibitem{fisher1936use}
R.~A. Fisher, ``The use of multiple measurements in taxonomic problems,'' \emph{Annals of eugenics}, vol.~7, no.~2, pp. 179--188, 1936.

\bibitem{van2008visualizing}
L.~Van~der Maaten and G.~Hinton, ``Visualizing data using t-sne.'' \emph{Journal of machine learning research}, vol.~9, no.~11, 2008.

\bibitem{kingma2014adam}
D.~P. Kingma and J.~Ba, ``Adam: A method for stochastic optimization,'' \emph{arXiv preprint arXiv:1412.6980}, 2014.

\bibitem{Schuld19evaluating}
\BIBentryALTinterwordspacing
M.~Schuld, V.~Bergholm, C.~Gogolin, J.~Izaac, and N.~Killoran, ``Evaluating analytic gradients on quantum hardware,'' \emph{Phys. Rev. A}, vol.~99, p. 032331, Mar 2019. [Online]. Available: \url{https://link.aps.org/doi/10.1103/PhysRevA.99.032331}
\BIBentrySTDinterwordspacing

\end{thebibliography}

\newpage
\onecolumn
\appendices
\crefalias{section}{appendix}

\section{Calculating the spectrum from the computational tree}
\label{sec:appendix_spectrum}
\Cref{algo:nemkov_tree} gives us access to a decomposition of \(f_\theta(x)\) in terms of trigonometric polynomials
\begin{equation}
\label{eq:trig_poly}
	f_\theta(x) = \sum_{\substack{s\text{, } c \in \mathds{N}_0^d \\
						s'\text{, } c'\in \mathds{N}_0^w}}
	k_{s, c, s', c'} \prod_{j = 1}^d \sin(x_j)^{s_j} \cos(x_j)^{c_j} \prod_{k = 1}^w \sin(\theta_k)^{s'_k} \cos(\theta_k)^{c'_k}, \nonumber
\end{equation}
where \(w\) denotes the number of variational parameters and \(k_{s, c, s', c'}\) is a set of coefficients, which is non-zero only for finitely many values of \(s, c, s' \text{ and } c'\).
We want to bring this into the form of \cref{eq:final_fourier_decomposition} shown in the main text, which allowed us to compute the spectrum and read off the Fourier coefficients. First, apply the well-known trigonometric identities
\begin{align*}
	\sin(x) &= \frac{1}{2i}\left(e^{ix} - e^{-ix}\right) \\
	\cos(x) &= \frac{1}{2} \left(e^{ix} + e^{-ix}\right).
\end{align*}
Inserting those into \cref{eq:trig_poly} yields
\begin{align*}
		f_\theta(x) = &\sum_{\substack{s\text{, } c \in \mathds{N}_0^d \\
						s'\text{, } c'\in \mathds{N}_0^w}}
	k_{s, c, s', c'} 2^{-\sum_{j=1}^d (s_j + c_j)} (-i)^{\sum_{j=1}^d s_j} \prod_{j = 1}^d \left(e^{ix_j} - e^{-ix_j}\right)^{s_j} \left(e^{ix_j} + e^{-ix_j}\right)^{c_j} \prod_{k = 1}^w \sin(\theta_k)^{s'_k} \cos(\theta_k)^{c'_k}.
\end{align*}
Using the binomial theorem we can further expand
\begin{equation*}
   {\left(e^{ix_j} - e^{-ix_j}\right)}^{s_j} {\left(e^{ix_j} + e^{-ix_j}\right)}^{c_j} = \sum_{a = 0}^{s_j} \sum_{b=0}^{c_j} \binom{s_j}{a} \binom{c_j}{b} (-1)^{s_j - a} e^{i (2a + 2b - s_j - c_j)x_j},
\end{equation*}
which finally yields an expression of \(f_\theta(x)\) as a Fourier sum
\begin{equation}
f_\theta(x) = \sum_{\substack{s\text{, } c \in \mathds{N}_0^d \\
						s'\text{, } c'\in \mathds{N}_0^w}}
	k_{s, c, s', c'} 2^{-\sum_{j=1}^d (s_j + c_j)} (-i)^{\sum_{j=1}^d s_j} \prod_{k = 1}^w \sin(\theta_k)^{s'_k} \cos(\theta_k)^{c'_k}  \prod_{j = 1}^d \sum_{a = 0}^{s_j} \sum_{b=0}^{c_j} \binom{s_j}{a} \binom{c_j}{b} (-1)^{s_j - a} e^{i (2a + 2b - s_j - c_j)x_j}.
\end{equation}
The two innermost sums, running over the indices \(a\) and \(b\), contain multiple terms which belong to the same frequency component \(2a + 2b - s_j - c_j\).
We want to summarize these terms and re-write the inner sums such that they run over unique frequency components.
The sum of all terms that contribute to a specific frequency \(\omega\) is given by \cref{eq:combinatorial_coefficients_sum}, which we restate here:
\begin{equation}
        p(s, c, \omega) = \prod_{j = 1}^d \sum_{a = 0}^{s_j} \sum_{b=0}^{c_j} \binom{s_j}{a} \binom{c_j}{b} (-1)^{s_j - a} \delta^{\omega_j}_{2a + 2b - s_j - c_j}.
\end{equation}
The inner two sums can be re-written in terms of the Gaussian hypergeometric function 
\begin{equation}
    p(s, c, \omega) = \prod_{j=1}^d \begin{cases}
        (-1)^{s_j} \displaystyle\binom{c_j}{(s_j + c_j + \omega_j)/2} \prescript{}{2}{F_1}\left(-s_j, - \frac{s_j + c_j + \omega_j}{2}; \frac{2 - \omega_j - s_j + c_j}{2}; -1\right) \text{, } \\  \text{if } -c_j - s_j \leq \omega_j \leq c_j - s_j \\
        \\
        i^{s_j + c_j - \omega_j} \displaystyle\binom{s_j}{(\omega_j + s_j - c_j)/2} \prescript{}{2}{F_1}\left(\frac{\omega_j - s_j - c_j}{2}, -c_j; \frac{2 + \omega_j + s_j - c_j}{2}; -1\right)\text{, } \\ \text{if } c_j - s_j < \omega_j \leq c_j + s_j \\
        \\
        0 \text{, else},
    \end{cases}
\end{equation}
which means that \(p(s, c, \omega)\) can easily be evaluated by a computer program.
\newpage

\section{Calculating the covariance between the Fourier coefficients}
\label{sec:appendix_covariance}
Assuming the variational parameters \(\theta\) to be drawn from a uniform distribution, the covariance of any pair of coefficients \(c_{\omega_1}(\theta)\) and \(c_{\omega_2}(\theta)\)
\begin{equation}
    \label{eq:covariance}
    \text{Cov}\left(c_{\omega_1}, c_{\omega_2}\right) = \frac{1}{(2\pi)^w} \int_{{[-\pi, \pi]}^w} c_{\omega_1}(\theta) c^*_{\omega_2}(\theta) \mathrm{d}\theta - \frac{1}{(2\pi)^{(2w)}} \left( \int_{{[-\pi, \pi]}^w} c_{\omega_1}(\theta)\mathrm{d}\theta \right)\left(\int_{{[-\pi, \pi]}^w} c^*_{\omega_2}(\theta) \mathrm{d}\theta \right)
\end{equation}
can be calculated exactly, as the coefficients only depend on \(\theta\) through powers of sines and cosines.
First, consider the mean of a coefficient function
\begin{equation}
    \int_{{[-\pi, \pi]}^w} c_{\omega}(\theta)\mathrm{d}\theta = \sum_{\substack{s\text{, } c \in \mathds{N}_0^d \\
						s'\text{, } c'\in \mathds{N}_0^w}}
	k_{s, c, s', c'} 2^{-\sum_{j=1}^d (s_j + c_j)} (-i)^{\sum_{j=1}^d s_j} p(s, c, \omega) \prod_{k = 1}^w \int_{-\pi}^{\pi} \sin(\theta_k)^{s'_k} \cos(\theta_k)^{c'_k} \mathrm{d}\theta_k.
\end{equation}
The integrals over powers of sine and cosine evaluate to
\begin{equation}
    \label{eq:powers_of_trigs_integral}
    \int_{-\pi}^{\pi} \sin(\theta_k)^{s'_k} \cos(\theta_k)^{c'_k} \mathrm{d}\theta_k = \begin{cases} &\frac{\pi 2^{-s'_k-c'_k} s'_k! c'_k!}{\left(\frac{s'_k}{2}\right)!\left(\frac{c'_k}{2}\right)! \left(\frac{s'_k + c'_k}{2}\right)!} \text{, if } s'_k \text{ and } c'_k \text{ even } \\
    &0 \text{, else}.
    \end{cases}
\end{equation}
The mixed term in \cref{eq:covariance} is calculated analogously.

\end{document}